remote sensing

Article

# Site Assessment and Layout Optimization for Rooftop Solar Energy Generation in Worldview-3 Imagery

Zeyad Awwad [1,†], Abdulaziz Alharbi [2], Abdulelah H. Habib [2,*] and Olivier L. de Weck [1,†]

1   Massachusetts Institute of Technology, Cambridge, MA, United States of America 02139; zeyad@mit.edu (Z.A.); deweck@mit.edu (O.L.d.W.)
2   King Abdulaziz City for Science and Technology, Riyadh, Saudi Arabia 12354; aasaalharbi@kacst.edu.sa (A.A); ahhabib@kacst.edu.sa (A.H.)
*   Correspondence: ahhabib@kacst.edu.sa
†   Current address: Engineering Systems Laboratory, Department of Aeronautics and Astronautics, Massachusetts Institute of Technology, Cambridge, MA, United States of America 02139

**Abstract:** With the growth of residential rooftop PV adoption in recent decades, the problem of effective layout design has become increasingly important in recent years. Although a number of automated methods have been introduced, these tend to rely on simplifying assumptions and heuristics to improve computational tractability. We demonstrate a fully automated layout design pipeline that attempts to solve a more general formulation with greater geometric flexibility that accounts for shading losses. Our approach generates rooftop areas from satellite imagery and uses MINLP optimization to select panel positions, azimuth angles and tilt angles on an individual basis rather than imposing any predefined layouts. Our results demonstrate that shading plays a critical role in automated rooftop PV optimization and significantly changes the resulting layouts. Additionally, they suggest that, although several common heuristics are often effective, they may not be universally suitable due to complications resulting from geometric restrictions and shading losses. Finally, we evaluate a few specific heuristics from the literature and propose a potential new rule of thumb that may help improve rooftop solar energy potential when shading effects are considered.

## 1. Introduction
### 1.1. Background

According to the International Energy Agency (IEA), global photovoltaic energy generation capacity reached nearly 160 GWp in 2020 [1]. Recent growth has been largely driven by the accelerating adoption of rooftop PV, which grew to nearly 60 GWp in 2020. However, in practice, residential PV system design remains a largely manual process. A number of automated approaches have emerged in recent years, but due to the computational complexity of the full optimization problem, they typically rely on predefined layouts, such as parallel rows or nonlinear neglect terms, such as shading. This paper describes a fully automated approach that employs 0.31 m RGB Worldview-3 satellite imagery to identify rooftops and subsequently generate complex solar panel layouts with detailed energy estimates that dynamically account for shading between panels during the optimization process.

As research into nonlinear optimization of multi-azimuth layouts with shading remains limited, we anticipate that the insights gained through this research will significantly contribute to the literature related to automated PV layout design. The proposed approach detects rooftops automatically from high-resolution satellites using morphological image processing methods. We use rooftop geometries to define a set of candidate panel locations and compute their expected energy generation. Next, we run a detailed geometric optimization that considers panel shading interactions to more accurately estimate energy generation and profit in a net metering scenario. Our analysis suggests that including these features in the optimization plays a significant role in site assessment and layout design, and may assist in improving solar potential through insights into the role of building design and urban layouts.

*1.2. Literature Review*

A small but growing body of work has examined the automated design of rooftop PV systems. Several commercial tools can provide partially automated designs for rooftop PV systems based on simple predefined layout options, typically arranging panels in rows facing a single azimuth angle [2]. While these tools are undoubtedly valuable for aiding site analysis of a single rooftop, they often involve some degree of manual operation (such as drawing rooftop polygons and identifying obstacles) and are typically limited in the potential layouts they can generate without human intervention.

1.2.1. PV Layout Design

Over the past few years, several academic groups have published work describing the use of remote sensing data and geometric optimization methods to automate the design of sophisticated rooftop layouts based on satellite and aerial data. Zhong et al. used LIDAR data to detect and characterize rooftops and obstacles that interfere with panel placement [3]. Since their work primarily focused on sloped rooftops and panels mounted parallel to the roof surface, they presented a limited shading analysis from fixed sources such as the rooftop and surrounding structures (but not the interactions between nearby panels).

Similarly, recent work by Narjabadifam et al. [4] employed 3D models from Google Maps as a starting point for optimizing the layout of rooftop PV systems. Their work included limited shading analysis between panels, considering the effect between parallel rows of panels along the same azimuth angle. While this approach can be effective for relatively large rooftops with few (or no) obstructions, it may not be suitable for complex rooftop geometries with significant obstructions. Rooftops with these properties are challenging for predefined layouts due to the geometric limitations of single-azimuth designs.

Despite these recent advances, several significant gaps still exist in the literature on the automated design of rooftop PV systems. The majority of existing work focuses on North American and European regions, which tend to feature homes with sloped roofs and commercial buildings with large and relatively open flat rooftops. In regions such as India and the Middle East, residential buildings typically have flat rooftops with partial stories and numerous small obstructions such as air conditioning units and water tanks. These types of buildings are well suited to satellite image analysis since, aside from relative heights, most of their features are parallel to the image plane and do not require depth or slope information to analyze. They are also more difficult to optimize since they can support arbitrary azimuth and tilt angles, whereas sloped roofs typically restrict panels to be parallel to the roof surface.

Additionally, current methods in the literature do not provide dynamic shading analysis of complex panel layouts. Due to challenges in both implementation and computational tractability, the implementations we reviewed often relied on heuristics designed for simple layouts such as parallel rows (which greatly simplify shading analysis) or neglected shading from neighboring panels entirely.

1.2.2. Design Heuristics

In both manual and automated rooftop PV layout design, several recurring design heuristics and rules of thumb are used to improve baseline system performance for predefined layout templates. These recommended approaches generally serve to improve the desired property (e.g., individual panel generation, packing density) or reduce an undesirable property (e.g., shading between panels). However, to the best of our knowledge, these heuristics tend to be imposed on layout design methodologies, and they have not been shown to emerge naturally from optimization methods. Our study will evaluate whether these recommended design rules are consistent with the results of a full MINLP optimization.

One of the simplest and most universal recommendations is to start with azimuth and tilt angles that feature high energy generation potential [5]. For non-equatorial regions, this typically means equator-facing panels tilted approximately toward the sun during peak daylight hours. This rule becomes more complex in multi-panel systems since increasing the tilt angle also increases the area shaded by that panel. The resulting shading losses may outweigh any gains in baseline energy potential, so a simple greedy maximization may lead to poor total generation.

Another common approach is to use parallel row layouts with fixed spacing, which reduces shading losses and also provides easy maintenance access to all panels (a federal regulatory requirement in our region of study [6]). This layout is simple to implement and tends to generate effective results. Studies show a packing density of 0.7 (representing the ratio of total panel area to rooftop area) for simple spaced row layouts on typical American rooftops [5]. International studies have shown that the optimal spacing (and resulting density) depends on latitude since it should be selected to balance increased panel density with the loss in energy due to shading [7].

Additionally, in the broader context of general polygon packing problems, several methods rely on heuristics to improve packing density. A common heuristic is to align at least one edge of the packing object to one of the edges of the bounding shape (in our context, representing the solar panels and the rooftop polygon, respectively). This approach forms the basis of the "advancing front" packing algorithm [8] that only places new polygons in contact with the active front, representing the accessible edges of the existing polygons (either the bounding polygon or a previously placed object). Similarly, some geometric bin packing methods demonstrated that relying on mated edges tends to minimize wasted space and improve packing density [9]. However, heuristics that increase packing density need to be used carefully, since increasing panel density beyond the ideal spacing tends to diminish generation performance because of increased shading losses [7].

*1.3. Contribution*

The study described in this paper aims to address the gaps described above. Our optimization method uses a novel independent set formulation that does not impose predefined layouts and can generate multi-azimuth system designs to maximize energy generation after accounting for shading interactions. Additionally, we evaluate several heuristics and rules of thumb used in packing algorithms and automated rooftop PV layout design (such as single-azimuth equator-facing rows, and the relationship between panel and building orientation) and the potential role of urban planning in improving rooftop PV potential. The supporting results are generated from a fully automated method that relies on common high-resolution satellite imagery that targets flat rooftops with irregular geometries and obstacles, characteristic of buildings in the Middle East.

## 2. Materials and Methods

*2.1. Overview*

The optimization pipeline consists of several preprocessing steps that are used to define the rooftop geometry and candidate panels, along with the corresponding energy and shading profiles. These features are used to define a mixed-integer linear problem (MILP) or mixed-integer nonlinear problem (MINLP) model within Pyomo, a Python-based optimization API that provides a unified interface between many open-source and commercial solvers [10,11].

Aside from the user options, the primary data inputs are a 0.31 m resolution satellite image and corresponding weather files for the geographic region. During preprocessing, we extract the rooftop area and detect potential obstacles from the satellite image to define a polygon for the optimization area (see Section 2.3.1). The rooftop geometry is used to filter the possible candidate panels using an accept/reject procedure. We determine potential locations by overlaying dense grids of panels with different azimuth angles, tilt angles and grid shifts. We retain the panel locations that are fully enclosed within the rooftop and are sufficiently far from obstacles and rooftop boundaries (0.3 and 0.6 m, respectively, by default).

The weather files are used to generate baseline energy generation before shading for the given panel specifications, tilt options and azimuth angle options (see Section 2.4.1). This computes the maximum energy each individual panel can generate. We also determine the relevant pairwise relationships between the candidate panels (geometric intersections and a time-dependent shading matrix) needed in the objective function and constraints.

The computed features are used to define the decision variables of the optimization problem, the expression for the objective function and the constraints that require enforcement. The model inputs are defined using the Pyomo API and passed to a compatible optimization solver. We use the SBB solver within GAMS [12] to generate results that include panel shading, and the linear solver Gurobi to generate the results that neglect to account for shading [13].

2.1.1. Data and User Inputs

The primary data inputs to the pipeline defined in Figure 1 consist of a high-resolution satellite image and the corresponding weather file for a given region. We use 0.31-m Worldview-3 imagery to generate all rooftop geometries using the morphological image segmentation model described in Section 2.3.1. We obtained the weather from NREL's National Solar Radiation Database (NSRDB) [14], which provides simulated hourly radiation and weather data generated by the Physical Solar Model (PSM) Version 3. The weather files are compatible with the Python version (PySAM) [15] of the System Advisor Model [16], using the built-in PVWatts model to estimate the energy generation profile of a given panel over time.

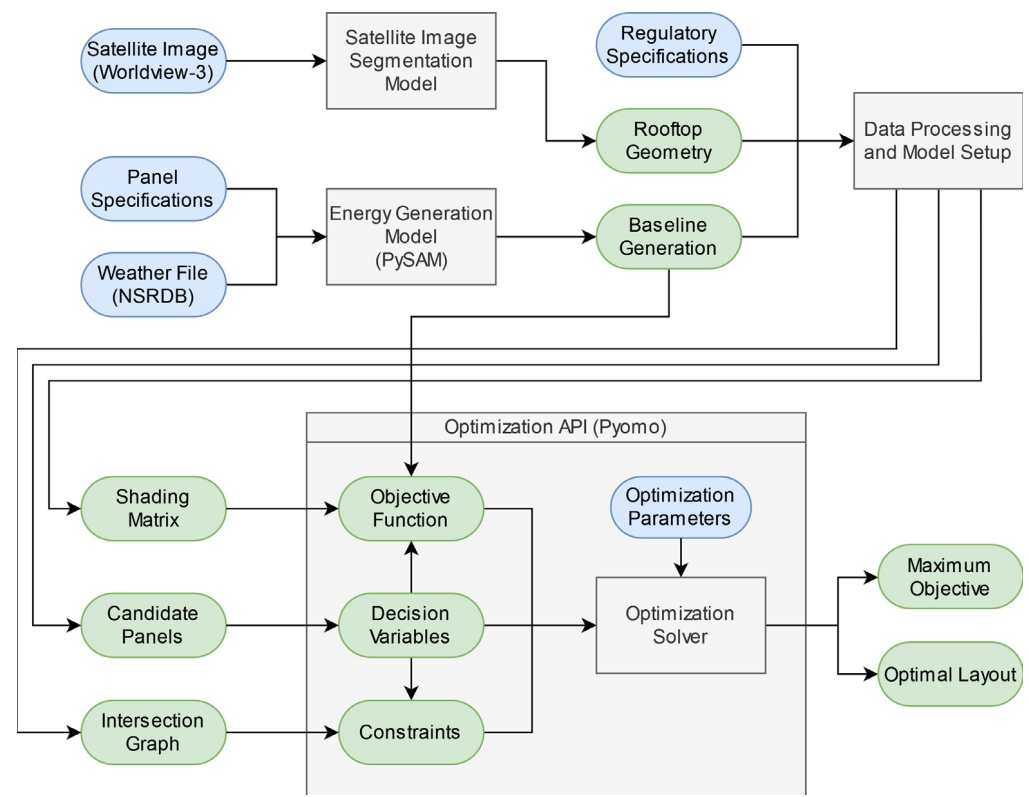

**Figure 1.** Full pipeline for rooftop solar panel optimization.

The full optimization pipeline also includes several user-defined variables that can be adjusted as needed. For our default panel specifications, we use 300 W panels with dimensions of 1.6 m × 1 m with an energy generation profile modeled by the PVWatts residential profile. The default panel placement options (detailed in Section 2.3.2) use eight azimuth angles (in 45° intervals), four tilt angles (in 10° intervals) and four possible grid shifts (in half-panel intervals). The setback and access buffers are 0.6 m by default, based on the minimum regulatory requirement for solar PV installations in Saudi Arabia [6].

*2.2. Optimization*

2.2.1. Model Formulation

To solve for the optimal panel placement, we frame the problem as an optimal independent set problem from graph theory. An independent set refers to any subset of nodes within a graph with no mutual connections—in other words, no edge exists between any pair of nodes in the set (see Figure 2 for examples). The maximum independent set problem is NP-complete [17], and our variation includes the added difficulty of maximizing a (potentially nonlinear) objective function rather than the number of nodes. The general form of this variation of the independent set can be represented as a constrained mixed-integer optimization problem in the form

$$\begin{aligned}\max_{\vec{x}} \quad & Objective(\vec{x}, \lambda_1, \ldots, \lambda_n) \\ \text{s.t.} \quad & x_i + x_j \leq 1 \quad \forall \quad (i,j) \in E \\ & x_i \in \{0, 1\}\end{aligned} \quad (1)$$

where the optimization method solves for the boolean vector $\vec{x}$ (representing whether or not a candidate panel is placed) that maximizes the given objective function. The constraint enforces the requirement that, for any edge between $(i, j)$ defined in the set of edges $E$, the solution can contain (at most) one of the two panels. The objective function may include additional parameters $\lambda_1, \ldots, \lambda_n$ that can be arbitrarily complex in principle; however, in practice, some objective functions can present additional challenges that must be compatible with the optimization solver employed.

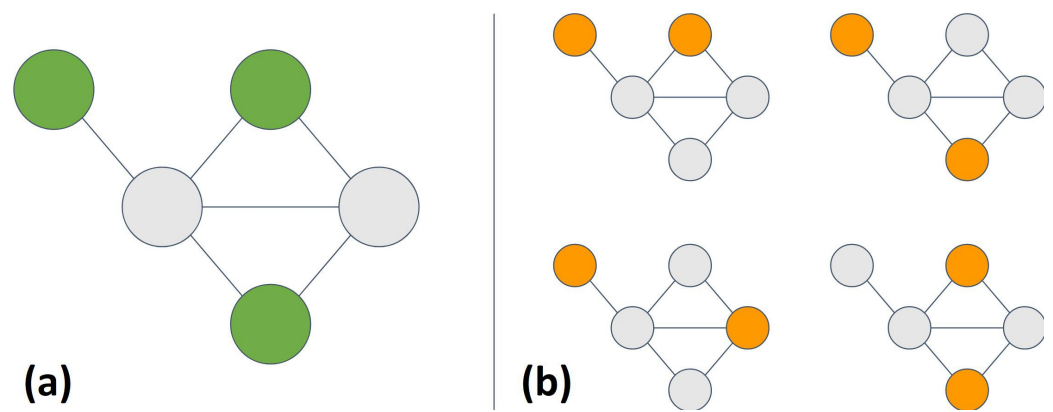

**Figure 2.** Examples of independent sets in a small graph showing (**a**) the maximum independent set and (**b**) all remaining independent sets with at least two elements.

To formulate panel selection and layout optimization as a maximum independent set problem, we construct an intersection graph on which the nodes represent candidate panels and the edges represent a pair of panels that cannot be simultaneously placed. We define an edge between any two panels that physically overlap or that would violate placement regulations—for example, regulations related to walkways and maintenance access. However, the conditions for defining an edge can be extended to represent any binary pairwise relationship that prohibits both panels from being present simultaneously.

An independent set, in this context, represents a valid combination of panels that respects physical and regulatory constraints. The objective function we use aims to maximize the return on investment of the layout by estimating the energy generation (after shading is accounted for) over a 20-year lifetime and subtracting the estimated installation cost. This could be extended to meet household demand profiles or improved to include additional installation details, but those factors are beyond the scope of this study.

2.2.2. Objective Functions

One of the most straightforward objective functions is the weighted sum of the form in Equation (2). This only requires a single additional parameter representing the panel weights, which can be selected to maximize different objectives. For example, using uniform weights maximizes the number of panels, while using the lifetime energy generation per panel maximizes the total energy generation. Similarly, by subtracting the total cost from the lifetime energy generation, this representation can also maximize a simple estimate of net profit when shading effects are neglected:

$$Objective(\vec{x}, \vec{c}) = \sum_{i=0}^{N} c_i x_i \tag{2}$$

This formulation is compatible with mixed integer linear programming (MILP) optimization solvers, which are widespread and relatively efficient when compared to nonlinear formulations. We use this formulation as a baseline to study the impact of shading losses in Section 3.4, where $\vec{c}$ is the estimated lifetime generation (neglecting shadows) minus the panel cost. However, treating each panel's energy generation independently fails to consider how panels may cast shade on each other and significantly diminish the energy generation of the final system [18].

We obtain a more realistic estimate of energy generation by incorporating a time-dependent nonlinear interaction term into the objective function, defined in Equation (3). For every panel $i$, we compute the total area shaded by all other panels $j$ at a specific time $k$. This is used to define a shading coefficient, between 0 and 1, which is multiplied by the baseline energy generation to estimate the total energy generation potential when shading effects are included.

This formulation, which involves a nonlinear objective function and a number of additional parameters, can be written as:

$$\max_{\vec{x}} \quad \sum_{i=0}^{N} x_i(-C_i + T \sum_{k=0}^{K} G_i(k)[1 - min(1, \sum_{j}^{N} x_j \cdot S_{ij}(k))])$$

$$\text{s.t.} \quad x_i + x_j \leq 1 \quad \forall \ (i,j) \in E$$

$$x_i, x_j \in \{0, 1\} \tag{3}$$

where

- $x_i$ = A boolean decision variable that represents whether candidate panel $i$ is selected in the solution (1 indicates that a candidate panel was selected, 0 if it is not);
- $N$ = The total number of candidate panels (decision variables) defined in this problem;
- $K$ = The total number of unique date-time combinations considered in the energy and shading calculations;
- $E$ = The set of edges in the graph representing all incompatible pairs of panels;
- $G_i(k)$ = The energy generated by panel $i$ at time $k$ (without shading), scaled such that the sum over the k time samples equals the annual generation;
- $S_{ij}(k)$ = The fractional area of panel $i$ that would be shaded by panel $j$ at time $k$;

$C_i$ = The marginal cost of adding panel $i$ to the layout;

$T$ = A scaling and conversion factor that, when multiplied by the annual generation by panel $i$, gives the corresponding estimated lifetime financial value.

In this formulation, the shading from all other selected panels $j \neq i$ onto panel $i$ at time $k$ equals the sum of all shading contributions (representing a worst-case shading scenario). Since the energy produced cannot be negative, the total shading is capped at 1, indicating that the panel is fully shaded.

The off-diagonal terms $S_{ij}$ (where $i$) represent the shading from panel $j$ onto panel $i$. The diagonal terms $S_{ii}$ are the baseline shading of panel $i$, representing fixed sources of shading on $i$ that are not affected by any decision variables, such as nearby objects or previously placed panels in neighboring regions (which are not included in $\vec{x}$).

This formulation is nonlinear and computationally demanding; as such, it is most effectively solved with specialized high-performance MINLP solvers. Smaller problems can also be recast into a form compatible with linear/quadratic solvers, such as Gurobi, by introducing additional variables and constraints. The shading coefficient can be represented by a piecewise linear variable (to bound the coefficient between 0 and 1) with additional constraints to determine the sum over all panels.

The formulation could be further generalized to incorporate more complex real-world factors. For example, the representation in Equation (4) uses a time-dependent tariff rate by replacing $T$ with $T(k)$ and absorbing it into the summation over $K$. This form can be used to incorporate dynamic pricing in regions where electricity prices vary based on the time of day. Additionally, since the true cost of solar installation depends on multiple nonlinear factors (including shared resources such as inverters and wiring), the constant panel cost $C_{panel}$ can be replaced by a cost function $C_i(\vec{x})$ that calculates the cost of adding a specific panel to a specific layout:

$$\max_{\vec{x}} \sum_{i=0}^{N} x_i(-C_i(\vec{x})) + \sum_{k=0}^{K} T(k)G_i(k)[1 - min(1, \sum_{j}^{N} S_{ij}(k)x_j)]$$

$$\text{s.t.} \quad x_i + x_j \leq 1 \quad \forall \quad (i,j) \in E$$
$$x_i, x_j \in \{0, 1\}$$

(4)

However, for the purposes of the current study, we use a fixed cost and tariff rate. This allows us to more directly evaluate the geographic and geometric characteristics of the problem without the added complexity and variability caused by incorporating time-varying local policies.

*2.3. Site Analysis and Preprocessing*

2.3.1. Satellite Image Segmentation for Rooftop Detection

We identify buildings by detecting and processing edges obtained by the morphological gradient algorithm [19]. This approach uses the difference between the dilation and erosion of a given image $I$ using a provided kernel $S$. The kernel is a small image patch used to define the neighborhood around each pixel using an image convolution. The image dilation ($\oplus$) and image erosion ($\ominus$) operations map each pixel to the brightest and darkest values within the kernel, respectively:

$$G = I \oplus S - I \ominus S \qquad (5)$$

The resulting image $G$ is defined by the difference between the brightest and darkest pixels within the neighborhood of each pixel, as defined by the kernel $S$. We found the morphological gradient to be especially well suited for building applications since it usually produced complete boundaries that fully enclosed objects. While alternatives such as the derivative-based Canny [20] and Sobel [21] detection methods are widely used for edge detection, they often produce disconnected sections of an edge with pixel-scale gaps that are not as well suited to object segmentation.

A closed boundary is critical for this approach since we detect objects using the flood-filling algorithm [19]. This algorithm takes a binary image and fills any holes, which are defined as regions that are enclosed within a complete internal boundary and disconnected from the outer boundary of the image. The input image is the maximum morphological gradient across all color bands, using a circular kernel with a radius of 3 pixels. The resulting image identifies all regions enclosed by a sharp change in pixel value in at least one color band, which (in our context) typically indicates different objects.

The results of the morphological gradient and flood filling algorithm are shown in Figure 3. When applied to high-resolution satellite images in a dense urban setting, these objects are most often buildings, vehicles, and road infrastructure. We distinguish between these objects using their geometric properties: vehicles can be filtered by their small size, and road infrastructure can be filtered by their elongated shapes [22]. We compute this by measuring the ratio of the maximum and minimum Feret diameters, which measure the largest and smallest possible separation between parallel lines that bound the object.

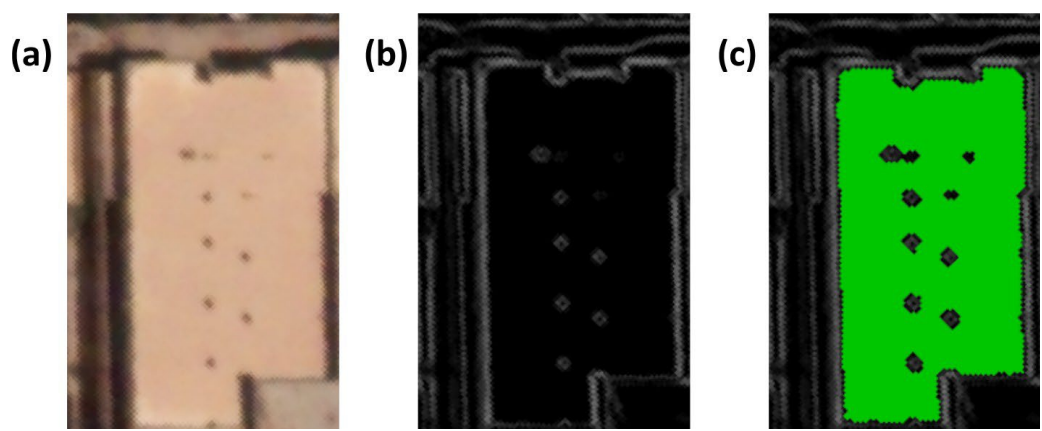

**Figure 3.** An example of rooftop detection showing (**a**) a zoomed-in region of the RGB input image; (**b**) morphological edge detection; and (**c**) the detected rooftop polygon with obstacles removed.

Optionally, to reduce misclassification and image segmentation errors, additional operations can be used to remove shadows and vegetation. Shadows are detected using the morphological shadow index (MSI) from Huang et al. [23], which identifies linear morphological operators that are significantly darker than their surroundings. We also filter vegetation using NDVI, which identifies surfaces that absorb most red light but reflect most near-infrared light [24].

2.3.2. Candidate Panel Generation

To generate the candidate panels that correspond to the set of decision variables $\vec{x}$, we must define their location and select one option for each of the three configuration parameters (azimuth, tilt, and shift). Using our default settings (shown in Figure 4), we generate a grid for each of the $8 \times 4 \times 4$ possible parameter options, resulting in 128 distinct configurations which will each be used to define a rectangular grid of panel positions (see Figure 5). Although increasing the parameter options can improve the resolution of the state space, the combinatorial nature of these parameter choices means that increasing the options quickly becomes prohibitively expensive during optimization.

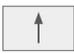

Figure 4. The default set of panel configuration options consisting of 8 possible azimuth angles, 4 possible tilt angles and 4 possible shift options.

The azimuth angles and grid shifts are used to define a grid of panels, which is overlaid on the detected region (see Figure 3(c) for an example) and then filtered to generate one subset of candidate panels. We require any candidate panels to satisfy two key requirements: they must be fully contained within the region, and they must comply with the minimum setback requirements. Regulations in Saudi Arabia require that all solar panels are a minimum of 0.6 m away from the edge of the rooftop and from any objects, which is implemented by applying a two-pixel buffer (corresponding to 0.61 m) around all internal and external boundaries in the rooftop geometry. Examples of the generated panel grids, which are combined to generate all candidate panels, are showcased in Figure 5.

Each of these candidate panels is assigned a baseline energy generation, as defined in Section 2.4.1, based on its azimuth and tilt angle. Additionally, for all pairs of panels included in a set of decision variables $\vec{x}$, we compute the pairwise shading terms $S_{ij}(k)$ and $S_{ji}(k)$ as defined in Sections 2.4.2 and 2.4.3.

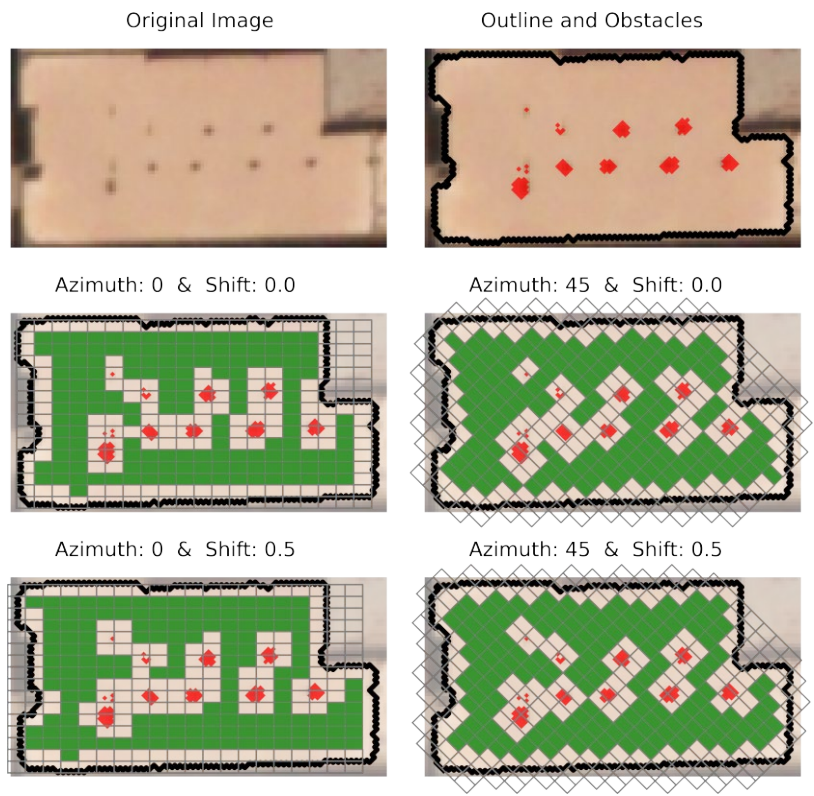

**Figure 5.** Examples of different panel grids (gray) overlaid on a detected building polygon (black) with obstacles (red). The panels that satisfy all placement requirements (shown in green) are viable candidate panels and are included as decision variables.

*2.4. Energy Estimation*

2.4.1. Baseline Energy Generation

Each candidate panel has a defined photovoltaic module, azimuth angle (which defines the compass direction of the surface normal), and tilt angle relative to a flat ground surface. We provide these parameters and the corresponding weather file into PySAM to estimate the baseline energy generation for every snapshot time. We used one year of hourly weather data for all of the locations in this study. These results provide the baseline energy generation for an unshaded panel for different times of day and different days of the year. These are combined with the shaded area of the panel to compute the final energy generation.

To incorporate the baseline energy generation profile into the objective function, we generate a small representative subset of the year by choosing one day per month with approximately even spacing across the year (the exact day is arbitrary, and our results used the 14th day of each month) and only considering major daylight hours (6:00 a.m. to 8:00 p.m. in the local time zone). These combinations provide a total of $K = 168$ time samples during significant daylight hours, which are scaled by 365/12 to recover the approximate annual energy generation. Since they are distributed approximately evenly across all seasons, they reasonably characterize the different shading conditions experienced in a typical year.

We chose to select one day per month because it provided a reasonable trade-off between accuracy and computational time. This choice reduced the number of terms from $K = 8760$ to $K = 168$, which is especially important for reducing the memory size and computational cost of $S_{ij}(k)$ as well as the objective function that must be computed during each step of the optimization. We found that this approximation produced annual generation estimates within 5% of the annual average under pairwise shading conditions.

2.4.2. Panel Shading

Photovoltaic panels are highly sensitive to shading in multiple respects. The most immediate effect is the reduction in energy generation, which can have a complicated nonlinear relationship with the area being shaded [18]. In the long term, nonuniform energy generation (both among different solar cells within a panel, and different solar panels connected to the same inverter) can significantly accelerate panel degradation. Although partial shading is difficult to avoid entirely, we reduce its impact by penalizing energy losses due to shading in the objective function.

We model the shading caused by panels using the shadow volume algorithm, one of the oldest and most common techniques for 3D shadow rendering [25]. This creates 3D volumes to represent shadows cast by objects and separate 3D space into two regions: all surfaces inside the volume are shaded, and all surfaces outside are not shaded by this object. In the case of a rectangular photovoltaic panel, the shadow volume takes the form of an oblique parallelepiped extending from the panel vertices on the surface ($P_1$) in the direction of the unit vector $\vec{s}$ representing the direction of sunlight (the azimuth and elevation angles) at that geographic location and the given time and date.

To compute the resulting shading on panel $P_2$ due to $P_1$, we need to determine the intersection of the shadow volume of $P_1$ with the surface of panel $P_2$. This is computed in two stages, using the geometric setup shown in Figure 6. The first determines the intersection of the shadow volume of $P_1$ on the infinite surface plane defined by $P_2$. If such an intersection exists, we proceed to the next step to determine how much of the panel is shaded (otherwise, there is no shading by $P_1$ on $P_2$ at this location at this time).

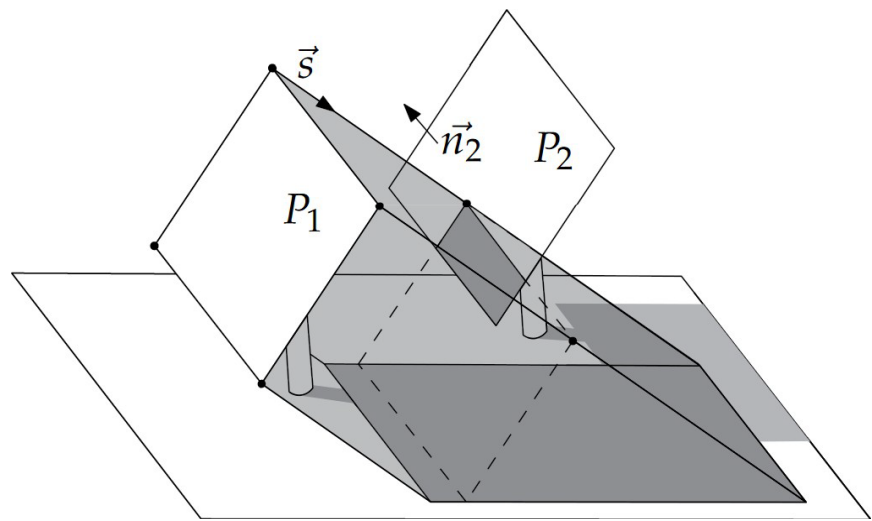

**Figure 6.** An example of the shading between two nearby panels at a single point in time, showing the shadow volume as a 3D parallelepiped and the shaded region as a dark gray quadrilateral on panel $P_2$.

If an intersection exists between the 3D shadow volume of $P_1$ and the 2D infinite plane containing $P_2$, there is a possibility that $P_2$ experiences shading due to $P_1$. The final step is to determine the intersection within the boundaries panel $P_2$, which we accomplish using a change of basis to the intersection plane to reduce the problem to 2 dimensions. This allows us to obtain the final shaded area by computing the 2D intersection of two polygons: one representing the panel $P_2$, and the other representing the intersection of the shadow volume of $P_1$ with the infinite plane of $P_2$.

2.4.3. Shadow Matrix

The shaded area computed using the methodology defined in Section 2.4.2, divided by the total area of the panel, represents the fractional shaded area on $P_2$ due to $P_1$. This is a fractional value between 0 and 1 (e.g., $S_{ij}(k)$ = 0.25 represents 25% shading by area), is approximately proportional to the energy loss and is used as a scaling factor for the total energy generated by that panel. Before running the optimization, we precompute all pairwise shading terms and store them in the shadow matrix element $S_{ij}(k)$ (where $i$ and $j$ are the indices of panels $P_2$ and $P_1$, respectively, and $k$ corresponds to the date–time pair being evaluated). This structure allows us to incorporate accurate pairwise shading in 3D space between two panels with arbitrary azimuth and tilt angles.

Additionally, this representation allows us to incorporate fixed sources of shading that are not affected by the decision variables. This is stored in the matrix diagonal $S_{ii}(k)$, and acts as a baseline shading term that always affects panel $i$. In principle, this can be used to represent any fixed sources of shading such as nearby walls, neighboring buildings and other rooftop objects. In our implementation, we use it to account for shading from previously placed panels in nearby regions, which contributes additional shading that can affect subsequent layouts.

One limitation of this implementation is that each pairwise shading term is computed individually, and shading from multiple panels is added linearly to obtain the total shaded area. Computing the exact combined shading from all panels would be too computationally prohibitive to include in the objective function. This represents the worst-case scenario where all shaded areas are non-overlapping, which may overestimate the total shaded area in cases of heavy shading from multiple sources.

*2.5. Implementation and Performance*

2.5.1. Sequential Optimization

Due to the computational complexity of MINLP optimization, and the highly combinatorial nature of the candidate panel generation defined in Section 2.3.2, solving a full rooftop in a single optimization may require prohibitively large amounts of RAM. In such cases, we use polygon decomposition to segment the rooftop into multiple regions that can be optimized sequentially. By limiting the maximum number of candidate panels in each segment (600 by default), we can optimize a rooftop in batches that place a hard limit on the size of the individual optimization runs.

To define the initial polygon segments, we use a network-based community detection approach to identify regions with clusters of high mutual visibility. In computational geometry, a pair of 2D points are visible to one another if a straight line between them does not intersect with any polygon boundaries [26]. We use this to define a visibility network, where nodes represent points along the perimeter and edges identify pairs of points that are visible to one another. The community detection method we employed was the random walk-based Walktrap algorithm [27] in Python's igraph library [28], which uses random walks to identify communities in large sparse graphs.

The resulting regions are typically separated by obstructions or bottlenecks in the polygon, since these would obstruct visibility between two adjacent regions. If any of these initial segments still contain more candidate polygons than the user-defined limit, they are further segmented through iterative bisection. After identifying the minimum rotated bounding box, we split the polygon into two using a bisection connecting the midpoints of the two long sides of the rectangle. This bisection procedure is repeated for any remaining polygon segments that exceed the candidate panel limit. An example of the initial Walktrap-based segmentation, and two different cases of recursive bisection, is shown in Figure 7.

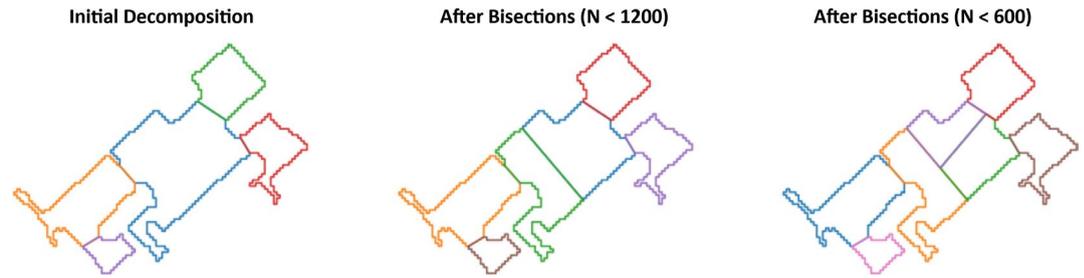

**Figure 7.** An example of rooftop polygon segmentation using the Walktrap-based decomposition (left) and two different bisections that restrict regions to a maximum number *N* of candidate panels.

To perform sequential optimization while accounting for shading interactions between regions, we treat all previously placed panels as fixed sources of shading for subsequent sets of candidate panels (as defined in Section 2.4.3). The pseudocode for our sequential optimization scheme is detailed in Algorithm 1. We use multiple sweeps along the sequence (2 by default) to allow earlier placements to be improved in response to shading from panels that were placed after. Although this approach will not generally give the same solution as a single large optimization, it enables greater scalability of MINLP geometric optimization since it can support arbitrarily large polygons while limiting maximum RAM usage.

---

**Algorithm 1** Pseudocode to implement the sequential optimization scheme

---

Subdivide rooftop geometry into regions $R_1, \ldots, R_N$
Assign every panel to the region $R_n$ containing its centroid
**for** iteration in 1:$N_{iterations}$ **do**
  **for** $n$ in 1:$N_{regions}$ **do**
    Clear subset of panels from solution by setting $x[i] = 0$ for every panel $i \in R_n$
    Compute shading on every $i \in R_n$ due to every panel $j \notin R_n$ that satisfies
      $x[j] = 1$ and assign the total fractional shaded area to $S_{ii}(k)$
    Solve Equation 3 for the subset of panels $x_n$ which contains only panels in $R_n$
    Update solution $x[i]$ for every panel $i \in R_n$ by copying the solution for subset $x_n$
  **end for**
**end for**

---

2.5.2. Hardware Specifications

The shading analysis and optimization represent the bulk of the running time and were executed on a desktop PC equipped with an Intel Core i7 12700k CPU (base clock: 3.6 GHz, boost clock: up to 5 GHz) and 128 GB of DDR4 RAM. Due to limitations on the RAM usage by the optimization solver, we used batch sizes of 600 candidate panels or less by default for the full MINLP problem in Equation (3). The running time for each building was highly variable, increasing with the total number of candidate panels, but a typical rooftop in our data set would complete the shadow matrix calculation and the layout optimization in a total of 10 to 20 min.

## 3. Results

In this section, we will showcase the results of our method in the three regions of study (representing different latitudes) with a particular focus on the distribution of tilt and azimuth angles in the optimized solutions. Additionally, we compare the results between the MINLP formulation, which includes shading, with the simpler MILP formulation that does not account for shading losses. Our MINLP formulation was validated against a standard spaced-row layout generated from the same candidate panels, which was implemented by identifying the largest set of parallel spaced rows that can be selected for each panel configuration (i.e., the combination of azimuth and tilt angles).

Our results found that our MINLP solutions were able to fit more panels and generate more energy than the best-performing set of parallel spaced rows. The performance gap was highly variable depending on the rooftop geometry and the presence of obstructions, which can diminish the performance of simple predefined layouts. For small rooftops with many obstructions, parallel row layouts tended to perform poorly and our MINLP method was able to fit 79% more panels and generate 76% more energy on average. For larger and more open rooftops, which can more effectively accommodate parallel rows, the solutions from the MINLP formulation had 23% more panels and generated 20% more energy on average.

*3.1. Regions of Study*

To extract rooftop geometries, we used 0.31 m Worldview-3 satellite images of a typical residential neighborhood in Riyadh, Saudi Arabia. These geometries can be artificially reassigned to any latitude, as long as the corresponding weather file is available. By using the same rooftop geometries in different regions, we can isolate the impact of generation potential and shading losses on the generated layouts.

In addition to our initial study region of Riyadh, Saudi Arabia (approximately 25° latitude, 48° longitude), we performed the same analysis in central Washington state (approximately 47° latitude and −120° longitude) and northern Brazil (approximately 1° latitude, −54° longitude). These regions were selected due to their weather history and nearly even spacing across latitudes. Central Washington is relatively dry and has nearly the same peak solar energy potential as Riyadh, making it well-suited for comparative analysis. Due to the limited availability of compatible weather data at lower latitudes, we selected this location in Brazil even though it exhibits significantly lower solar potential overall (approximately 35% lower annual generation for the highest-performing panels).

To explore the role of urban planning and solar efficiency, we also vary the rotation angle of each rooftop geometry (as shown in Figure 8). We define the orientation of the principal axis as the longest side of the minimum rotated bounding box enclosing the polygon. Rotation angles are relative to the equator, where a rotation angle of 0° aligns the longest side of the bounding box to the horizontal axis (maximizing the width of the rooftop). We analyze rotations between 0° and 90°, in intervals of 22.5°, to evaluate how urban layouts may impact rooftop solar potential.

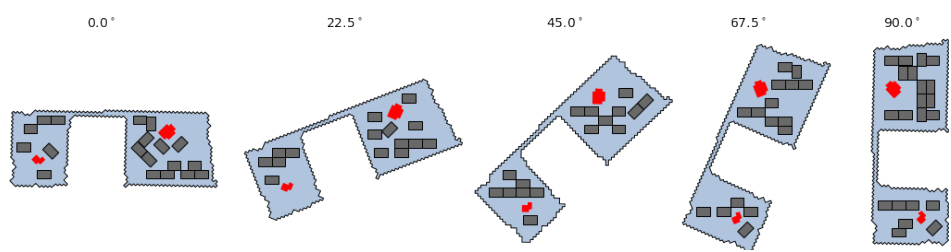

**Figure 8.** An example of a detected building (with obstacles) rotated to the five default alignments, and the resulting panel placements.

Our primary variables relating to panel configuration are the panel azimuth angle (where 0° is north-facing and 90° is east-facing) and the tilt angle (where 0° is parallel to the ground and 90° is perpendicular). We will explore potentially general design characteristics by examining statistical differences for different regions and rotation angles. The performance of each azimuth and tilt angle, measured as the 20-year return on investment under unshaded conditions, is shown for each region in Figure 9. The vertical axis shows the net profit divided by the cost, such that 0 represents breaking even, and 0.5 represents a profit that equals 50% of the panel cost. The resulting ROI values are consistent with typical payback periods of approximately 8–14 years for rooftop PV systems [29].

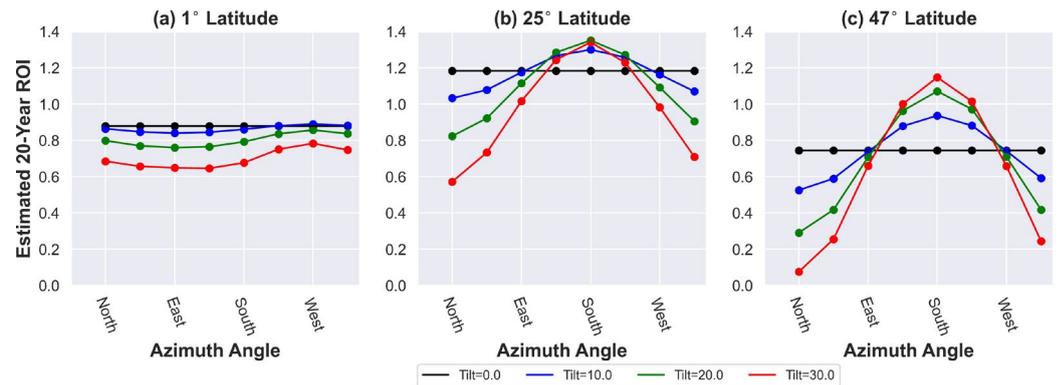

**Figure 9.** The 20-year return on investment (ROI) for a single panel (representing the net profit divided by the cost) based on PySAM generation profiles for the three different latitudes, computed for different panel tilts and azimuth angles using the default panel cost and tariff rate.

*3.2. Effect of Latitude on Panel Configurations*

The tilt angles in the optimized layouts (shown in Figure 10) are generally consistent with the highest-performing choices in Figure 9. For moderate and high latitudes, the highest-performing tilt angle represented approximately 70% of panels placed in the solution. Near the equator, the situation is slightly more complex since the two lowest tilt angles (0° and 10°) both achieve comparable performance: there is a slight preference for 10° tilt angles for partially or fully west-facing panels, and a preference for 0° tilt angles for all other azimuth angles. As a result, the solutions feature a mixture of the two tilt angles with a preference for 10° overall since they allow for slightly higher performance.

The remaining tilt angles occur in small quantities, which could occur for a variety of reasons. The rate of 10–15% (for moderate to high latitudes) may result from trying to avoid shading losses, since low tilts cause less shading and high tilts are less susceptible to shading. In specific cases, this may incentivize the use of lower-performing panels if they lead to reduced shading losses. Additionally, since MINLP problems have high computational complexity and solvers typically cannot guarantee global optimality, the obtained solution may reflect characteristics specific to a local optimum.

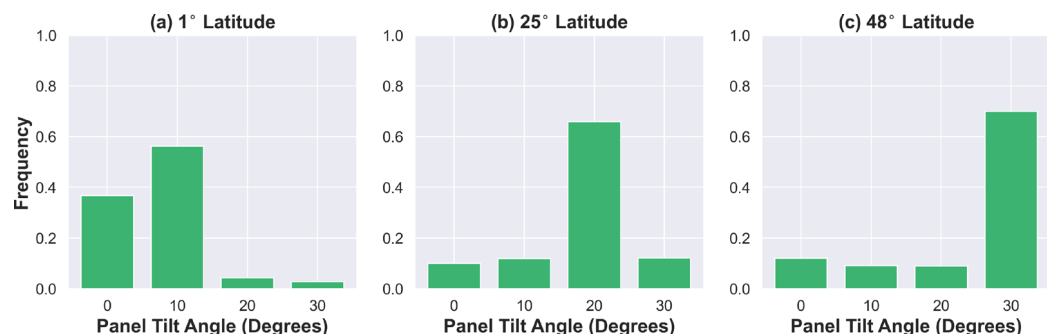

**Figure 10.** The frequency of selected tilt angles for solutions at different latitudes.

Similarly, looking at the distribution of azimuth angles in Figure 11 shows a strong preference for the highest-performing panel configurations shown in Figure 9. Away from the equator, approximately 70% of the panels are south-facing, and the remaining azimuth angles are present in small quantities. A notable exception is north-facing panels at 25° latitude, which appear more often than some higher ROI configurations. Since these conveniently fit behind south-facing panels without shading losses (for equal tilt angles), they can be used to fill in small gaps where panels with a higher individual ROI may suffer from higher shading losses or might not fit at all (due to spacing and access regulations).

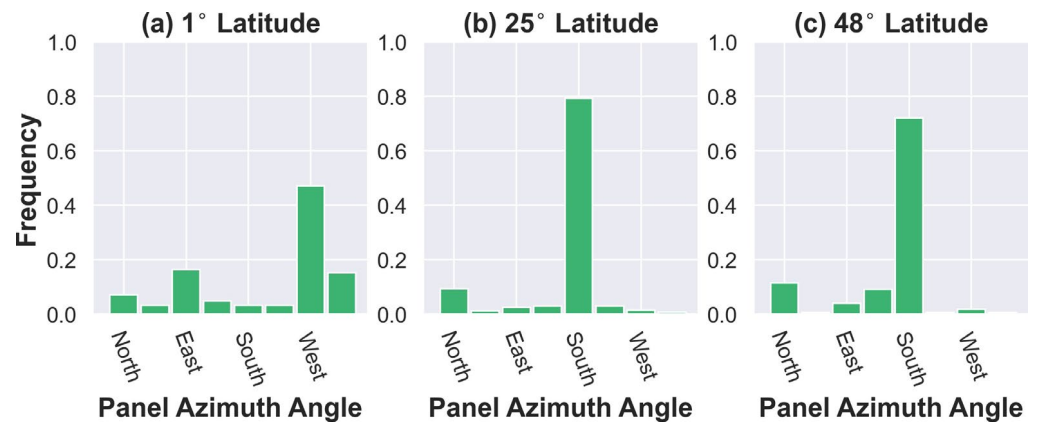

**Figure 11.** The frequency of selected azimuth angles for solutions at different latitudes.

Near the equator, at least based on the weather profile in the Amapa region of Brazil, there is a preference for west-facing panels consistent with the maximum ROI configuration shown in Figure 9 for tilt angles of 10°. As before, we see a second peak at the opposite azimuth angle of the main peak. Since east-facing panels can be placed directly behind west-facing ones and face little or no shading losses, they are a geometrically convenient way to fill gaps that would not permit or reward panels that have a higher baseline generation.

*3.3. Impact of Building Rotation*

To examine the effects that urban planning can have on rooftop solar energy potential (with a particular focus on parcel layout), we artificially rotate the building to align its principal axis to one of five rotation angles ranging from 0° (horizontal) to 90° (vertical). Figure 12 shows how the energy generation potential changes depending on the rotation angle for each latitude.

For medium and high latitudes, we can see that the 0° (horizontal) to 90° (vertical) rotation angles consistently outperform the rest, and, for all latitudes, the 45° angle consistently produces the least energy. The equatorial case is the only result where the 0° building exhibits notably weak performance, which may be a consequence of its unique shading properties. Since shading near the equator is almost entirely along the east-west axis, the horizontal alignment would be the most susceptible to shading losses, and a vertical or near-vertical alignment would be the least susceptible.

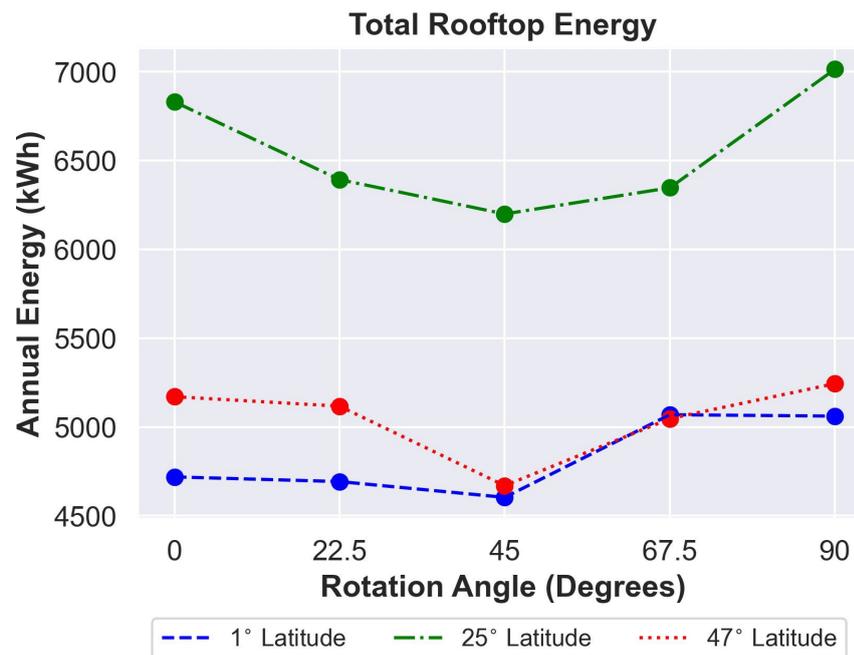

**Figure 12.** The total annual energy generated, as an average across all tested rooftops, for different degrees latitude, as a function of the building rotation angle.

In principle, this could be caused by a number of causes that may either be general to the problem or specific to the formulation. These include panel efficiencies, shading effects and discretization of the solution space (which can affect the maximum geometric capacity). Although we can not fully decouple these potential causes from the results of the optimization, we can obtain some insight by exploring the role of shading effects and how they impact the efficiency of resulting panel configurations.

*3.4. Impact of Shading Losses*

In general, the average shading losses tend to remain low (an average of 1–3% loss for all latitudes and rotation angles) even though individual panels in the solution sometimes exhibit shading losses as high as 20% (or even 30% at high latitudes). This suggests that, although significant shading losses do not necessarily eliminate a candidate panel from selection, they appear to be strongly disincentivized in nearly all cases and rarely selected by the optimization solver in practice.

We can examine the effect of shading on the optimization by comparing the packing density for the full optimization with shading, as defined in Equation (3), with a simplified variant that has all shading disabled, which can be reduced to the form of Equation (2) with an appropriate choice of the weights $C_i$. These result in a number of changes to the resulting layouts and panel characteristics. The resulting packing density for different rotation angles is shown in Figure 13.

When shading is included, the resulting layouts have the highest generation packing density when the buildings have a principal axis along or near the horizontal (0°) or vertical (90°) rotation angles. These building alignments represent the "wide" and "tall" orientations of a given rooftop, and can most easily support simple parallel row arrangements with low shading losses. The 45° alignment, however, consistently exhibits the lowest packing density for all latitudes when shading is included.

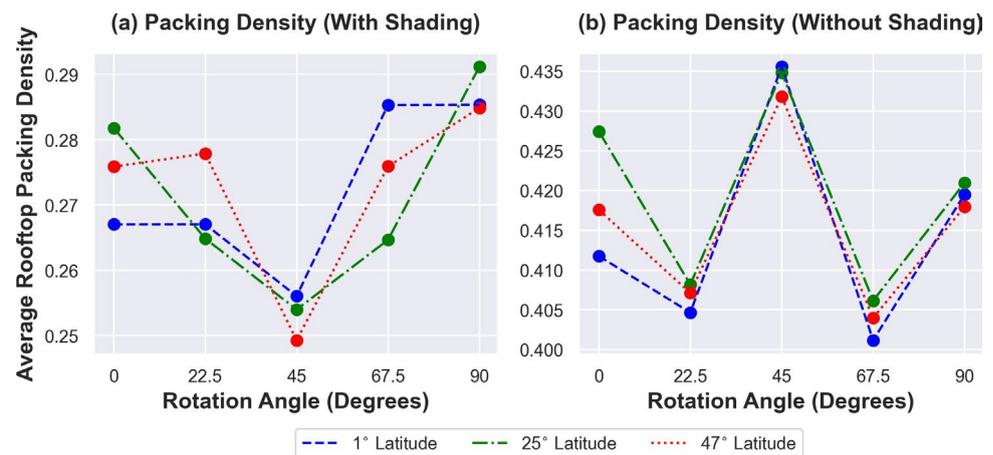

**Figure 13.** The packing density (ratio of panel area to rooftop area) of solutions to the shaded (**a**) and unshaded (**b**) optimization formulations.

In contrast, in the solutions to the optimization without shading, the 45° alignment consistently outperforms all other building rotation angles at every tested latitude. This result suggests that, unlike the 22.5° and 67.5° alignments, the 45° alignment tends to have an advantage in terms of geometric capacity. Since the 22.5° and 67.5° alignments are not parallel to any of the eight default panel azimuths, they consistently have the lowest packing density, but this disadvantage no longer exists when the panel azimuth options are increased to avoid that limitation (see Figure 14).

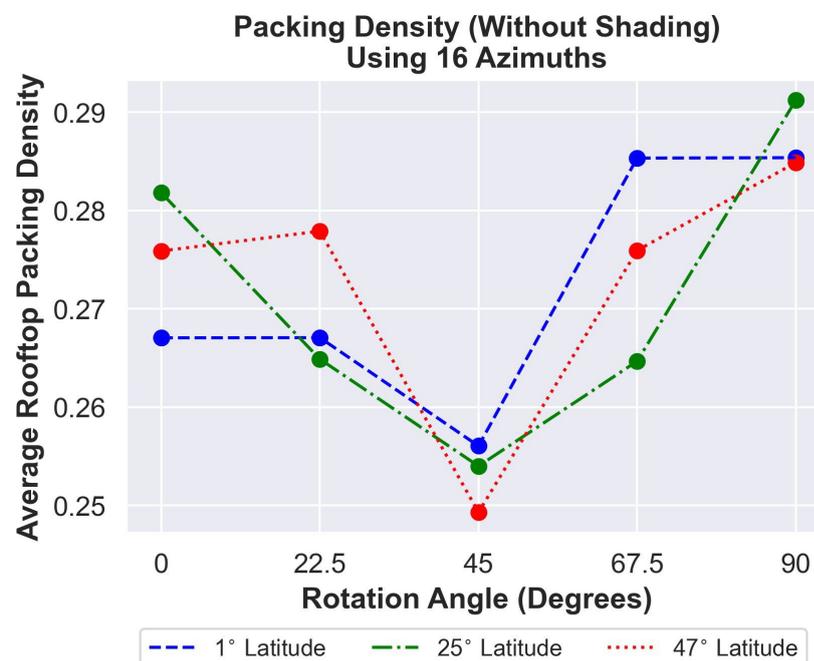

**Figure 14.** The average packing density, after doubling the possible azimuth angles to 16, for solutions to the unshaded optimization problem.

However, despite its high geometric capacity, the 45° alignment appears to be the least capable of effectively utilizing the available space when shading effects are included. Comparing the choice of azimuth angles (see Figure 15) for the two cases can help shed some light on the driving factors behind these differences. When the optimization is performed without shading, we can see that higher latitudes are significantly more likely to use a variety of azimuth angles to achieve a higher packing density. Although the individual panels have a lower average baseline energy generation, this is outweighed by the increase in panel count when interaction effects are neglected.

In combination, these results suggest that panel shading interactions play a crucial role in the design of effective rooftop PV layouts, and can have different effects depending on the latitude of the site. Without shading, the optimized solutions tend to prioritize high density over individual panel performance. When shading is included, the most effective layouts at higher latitudes are often dominated by a single azimuth angle with high individual panel generation (with a tendency to form spaced rows), using a mix of other panel configurations in small quantities to fill in gaps without substantially increasing the average shading loss.

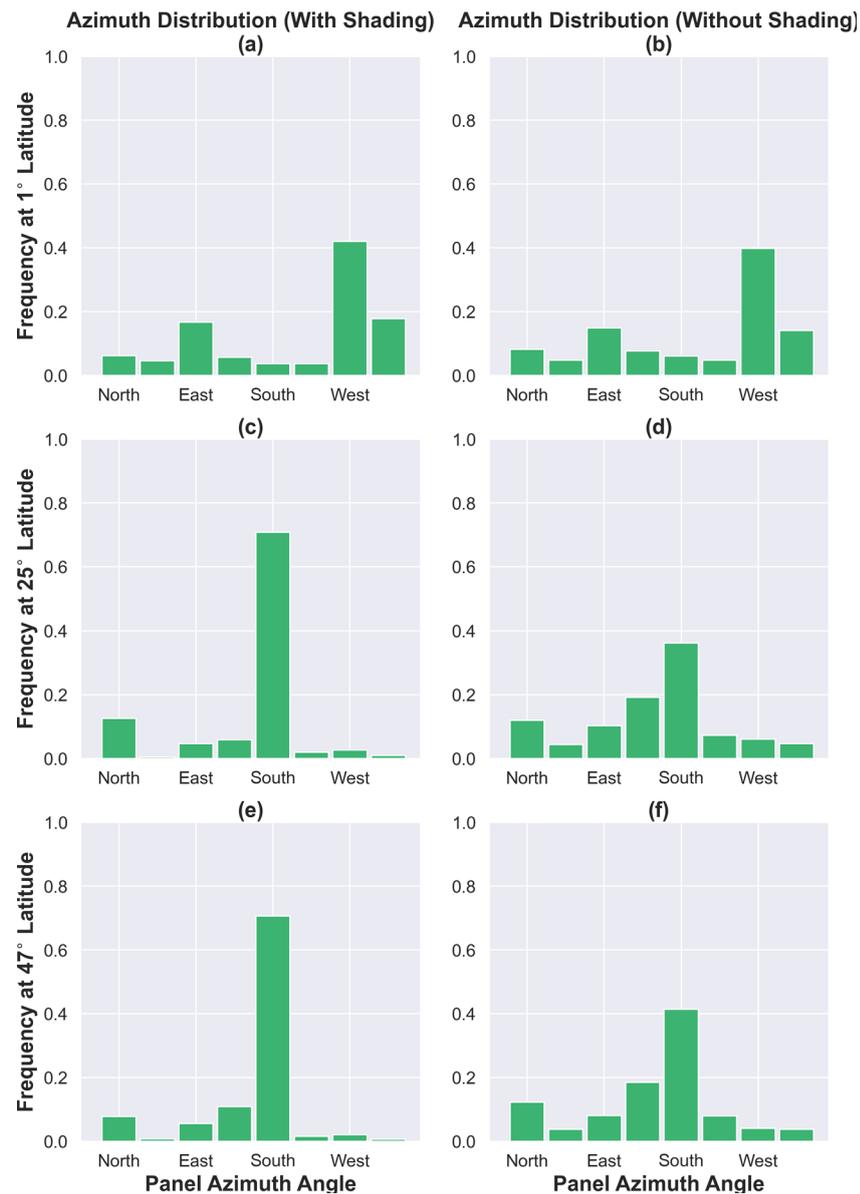

**Figure 15.** The distribution of panel azimuth angles in solutions to the shaded (left column) and unshaded (right column) optimization formulations at different latitudes. Results (a) and (b) compare the results with and without shading interactions for the weather profile near the equator. Similarly, (c) and (d) show a comparison at 25° latitude, and (e) and (f) show a comparison at 47° latitude

## 4. Discussion

The methodology presented in this paper demonstrates the potential of image processing and MINLP optimization methods to evaluate rooftop solar energy potential and layout design. The results suggest that shading interactions play a critical role in the design choices of optimized layouts, particularly at locations far from the equator. The results suggest that, although single-azimuth equator-facing row layouts are an effective rule of thumb for greater latitudes, overall performance improves by multi-azimuth layouts that primarily (but not exclusively) follow an equator-facing row structure.

In addition to using the generating PV layouts directly for building-level site assessment and energy projection, this methodology also suggests that urban planning and building geometry may play an important role in the performance of residential rooftop PV installations. In particular, building alignment (which is typically parallel to adjacent road and parcel boundaries) seems to affect layout performance significantly. This effect appears to be driven by shading interactions.

There are several potential opportunities for further research, encompassing improvements to the optimization and extensions to the analysis. Alternate formulations might be able to avoid the limitations caused by the discretized panel configurations in our approach. Additionally, further work is needed to establish the relationship between urban planning choices (particularly how they affect building geometry) and the efficacy of rooftop PV layouts at greater latitudes.

### 4.1. Residential PV Layout Design

The results of our optimization methodology demonstrate a few key characteristics (when shading effects are included) that suggest or verify general best practices for PV layout design. They suggest that the common approach of using single-azimuth layouts tends to be effective in cases where they are geometrically viable (e.g., solar farms and large, open rooftops) but can often be improved by allowing additional azimuth angles. Although these non-primary azimuth angles have lower baseline energy capacities, they increase the overall generation by filling in gaps in irregular geometries without substantially increasing shading losses.

However, comparing results across different latitudes suggests that these best practices are not universal and differ significantly between equatorial and non-equatorial regions. The average distribution of azimuth angles is dominated by equator-facing panels (approximately 70%) at greater latitudes, but only exhibits a modest preference (approximately 40%) towards the highest-performing azimuth angle for equatorial regions. In non-equatorial regions, the optimized solutions tend to feature single-azimuth rows, while solutions near the equator tend to result in less structured arrangements.

### 4.2. Building Geometry and Urban Planning

In addition to comparing results at different latitudes, altering the alignment of a building's principal axis (relative to the equator) allows us to examine the role that urban planning might play in renewable energy adoption. Since building alignments are typically parallel to road and parcel layouts, it may provide an opportunity for policymakers to help increase the solar generation potential of residential rooftops. Our results suggest that buildings that are primarily parallel or perpendicular to the equator are the most effective, with results that vary slightly by latitude.

At greater latitudes, the highest performing rooftops tended to be approximately parallel or perpendicular to the equator (0° or 90°). In contrast, the 45° alignment consistently produced the least energy and contained the fewest panels. Comparing this with the shadow-free optimization (see Figure 13) indicates that this is driven primarily by shading effects rather than geometric characteristics. Near the equator, however, shading is directed almost entirely along the east-west axis so the vertical and near-vertical alignments exhibit the best performance for the optimization with shading. Since the vertical alignment minimizes the width of the rooftop polygon, it appears to be the most effective layout to reduce shading effects.

Urban planners and policymakers might be able to increase residential rooftop PV adoption and generation potential by establishing guidelines for solar-friendly building designs. The results in this paper suggest that horizontally and/or vertically aligned buildings tend to produce the highest output, depending on the degree latitude, but only when shading effects are accounted for. Future research into the sensitivity to building alignment and other geometric characteristics could be of interest to urban planners, architects and government officials.

*4.3. Evaluating Common Design Heuristics*

Based on the results shown in Sections 3.2–3.4, the optimized layouts exhibit patterns that appear broadly consistent with the common design heuristics reviewed in Section 1.2.2. However, we see a sensitivity to both latitude and shading effects that should be taken into consideration when applying these heuristics. Additionally, we can see where our more general layout generation method breaks these patterns to improve performance.

The first heuristic described in Section 1.2.2, which encourages using equator-facing panels at a specific tilt angle, seems effective at higher latitudes but less significant near the equator. For latitudes 25° and 47°, around 70% of panels were south-facing in the final optimized layouts (typically at, or near, the highest performing tilt angle). Interestingly, this pattern was not prominent when we excluded shading interactions, suggesting that shading effects may be at least partially responsible for the efficacy of this heuristic. Without shading effects, the optimization method uses a more diverse mix of azimuth angles to prioritize high packing density instead.

The second heuristic, which recommends using spaced row layouts, is also consistent with the results when shading interactions are included. This partially follows from the previous case (the preference for equator-facing panels) but also relies on the ability of spaced rows to minimize shading interactions. When we exclude shading effects, the optimized layouts are much denser and seldom feature row structures. As before, this heuristic primarily matched the results at higher latitudes and did not appear to be significant near the equator.

In contrast, the third heuristic (to align panels to building geometry) has merit under certain conditions but appears to become considerably less useful when shading interactions are included, at least for the rooftops examined in this study. This design pattern seems to emerge only when we neglect shading losses (see Figure 13) and disappears when shading losses are included. This suggests that this heuristic is dominated by shading interactions in this context and may be ineffective for panel placement for flat residential rooftops of these sizes, even if these results support this heuristic for layout optimization without shading (which suggests it may be a good rule of thumb in purely geometric packing problems).

The first two heuristics (equator-facing panels and spaced-row layouts) seem to emerge naturally from shading interactions under general-purpose optimization, and are consistent with around 70% of the panels present in the optimized layouts. In contrast, the third heuristic (aligning panels to the building geometry) is only supported by the results when shading effects are neglected, suggesting that it is ineffective in this context even though it has been shown to be effective for other types of packing problems. Although the third heuristic allows for increased packing density, in this context, this appears to be counterproductive due to increased shading losses. These observations support using heuristics that prioritize high individual panel outputs and low shading losses, but not heuristics that increase packing density.

*4.4. Limitations and Future Work*

There are several built-in assumptions in the formulation that affect the solutions and should be kept in mind when interpreting these results. The spacing and access regulations in Saudi Arabia (such as a 60 cm minimum buffer in front of every panel) certainly contribute to the tendency towards spaced row structures. In regions without such restrictions, spaced rows would likely be outperformed by dense flat layouts without spacing at 0° tilt (since that would eliminate shading interactions). Additionally, the discrete panel configuration choices (azimuth, tilt and location) introduce discretization error and lead to a combinatorial explosion in the number of decision variables, which limits the parameter-space resolution.

Additionally, the solutions for nonlinear and non-convex objective functions are not guaranteed to be globally optimal using our solver, and, for large problems, such a guarantee is typically not computationally tractable. Further research into alternate formulations and solvers may help improve the performance of the final solutions, and comparative studies between different formulations may shed light on the specific properties suggested in our results.

## 5. Conclusions

Throughout this paper, we demonstrate the successful implementation of multi-azimuth PV layout optimization with shading interactions. The results suggest that a combination of azimuth angles often outperform single-azimuth designs for flat residential rooftops with obstructions. We also saw that shading losses played a large role in layout design, with markedly different effects near the equator compared to higher latitudes. Our results suggest that accounting for shading is critical for effective automated rooftop PV design.

Additionally, we evaluated several heuristics by identifying patterns in the optimized results (with and without shading). Our results also suggest that heuristics that improve packing density may not be useful when shading interactions are included. In contrast, heuristics that reduce shading losses (such as equator-facing panels and spaced rows) are consistent with optimized layouts and appear to emerge naturally in the context of shaded PV optimization. However, since these results are based on a limited case study, further research is needed to establish the efficacy of these heuristics.

Since there is limited existing research that uses MINLP for nonlinear PV layout optimization with nonlinear partial shading interactions, there are several potential avenues for future research. Different formulations may be more computationally efficient or generate higher-performance layouts. Alternatively, there may be more effective optimization solvers that can find better solutions for nonlinear, non-convex objectives.

**Author Contributions:** Conceptualization, Z.A., A.A. and A.H.H.; methodology, Z.A. and A.A.; software, Z.A. and A.A.; validation, Z.A., A.A. and A.H.H.; formal analysis, Z.A.; investigation, Z.A.; resources, Z.A.; data curation, Z.A.; writing—original draft preparation, Z.A. and A.A.; writing—review and editing, Z.A., A.A., A.H.H. and O.L.d.W.; visualization, Z.A.; supervision, A.H.H. and O.L.d.W.; project administration, A.H.H. and O.L.d.W.; funding acquisition, A.H.H. and O.L.d.W. All authors have read and agreed to the published version of the manuscript.

**Funding:** This research was funded by the King Abdulaziz City for Science and Technology (KACST).

**Data Availability Statement:** The two data sets used in this study were (1) Commercial Worldview-3 satellite data, which can be purchased from authorized vendors, and (2) weather data provided by NREL's National Solar Radiation Database (NSRDB), which can be accessed at https://nsrdb.nrel.gov/data-viewer (accessed on: 15 July 2022.

**Conflicts of Interest:** The authors declare no conflict of interest.